# Environmental force sensing helps robots traverse cluttered large obstacles using physical interaction

Qihan Xuan, Chen Li

Department of Mechanical Engineering, Johns Hopkins University

Corresponding author: [redacted], https://li.me.jhu.edu

## Abstract

**Many applications require robots to move through complex 3-D terrain with large obstacles, such as self-driving, search and rescue, and extraterrestrial exploration. Although robots are already excellent at avoiding sparse obstacles, they still struggle in traversing cluttered large obstacles. To make progress, we need to better understand how to use and control the physical interaction with obstacles to traverse them. Forest floor-dwelling cockroaches can use physical interaction to transition between different locomotor modes to traverse flexible, grass-like beams of a large range of stiffness. Inspired by this, here we studied whether and how environmental force sensing helps robots make active adjustments to traverse cluttered large obstacles. We developed a physics model and a simulation of a minimalistic robot capable of sensing environmental forces during traversal of beam obstacles. Then, we developed a force-feedback control strategy, which estimated beam stiffness from the sensed contact force using the physics model. Then in simulation we used the estimated stiffness to control the robot to either stay in or transition to the more favorable locomotor modes to traverse, depending on the estimated beam stiffness. When beams were stiff, force sensing induced the simulated robot to transition from a more costly pitch mode to a less costly roll mode, which helped the robot traverse with a higher success rate and less energy consumed. By contrast, if the robot simply pushed forward or always avoided obstacles, it would consume more energy, become stuck in front of beams, or even flip over. When the beams were flimsy, force sensing guided the robot to simply push across the beams. In addition, we demonstrated the robustness of beam stiffness estimation against body oscillations, randomness in oscillation, and uncertainty in the robot's position sensing. We also found that a shorter sensorimotor delay reduced mechanical energy cost of traversal.**

# 1. INTRODUCTION

Mobile robots are becoming increasingly prevalent in society. Many important tasks require robots to move through complex terrain with large obstacles (comparable to robot size or larger), such as furniture in household chores [1], buildings, vehicles, and pedestrians in self-driving [2], rubble in search and rescue [3,4], rocks and vegetation in forests [5], and boulders in extraterrestrial exploration [6]. A common strategy for moving in such environments is to avoid obstacles [7–9]. Sensors like cameras, radars, and LiDAR are used to create a geometric map of the surroundings, then a collision-free path is planned to maneuver around obstacles [10,11]. This geometry-based approach is effective in environments with sparse obstacles [1,2].

However, obstacle avoidance faces difficulties in environments with cluttered large obstacles. First, a collision-free path may simply not exist. Second, a collision-free path may be energetically too expensive or require a long execution because the robot must take a long detour to avoid obstacles. Furthermore, if not avoiding obstacles promptly, unexpected collisions and contact could damage the robot or cause it to to flip over [12,13]. To improve robot mobility in terrain with cluttered large obstacles, it is crucial to understand how to use and control physical interaction with enviroment (environmental affordance [14,15]). For example, understanding tire dynamics [16] and terramechanics [17,18] has helped wheeled vehicles and robots move on paved roads and off-road deformable ground, respectively. Understanding leg-ground interaction [19–21] has provided the foundation for legged robots to walk and run on flat surfaces with small unevenness. Recent research into leg interaction with granular media has facilitated the robotic design and control to improve legged and even (surprisingly) wheeled locomotion on deformable ground [22–27]. Similarly, understanding how to use and control the physical interaction between robots and obstacles will enhance their ability to move in complex 3-D terrain with cluttered large obstacles.

Terrestrial animals often use their body to make direct physical contact with large obstacles to traverse them [28,29]. Recent studies have shed light on how these animals traverse large obstacles through direct physical interaction. Through biological experiments [28,30,31], robophysical modeling [30,32–37], and theoretical and computational modelings [38–40], researchers have begun to understand how animals and robots use physical interaction to traverse complex 3-D terrain.

Our work is motivated by the exceptional capabilities of cockroaches in traversing complex 3-D environments and enabling robots to have similar capabilities. The rainforest-dwelling discoid cockroach, which is excellent at traversing cluttered large obstacles such as vegetation, foliage, crevices and rocks. A previous study discovered that it can traverse flexible, grass-like beam obstacles using and transitioning across different modes of locomotion [28]. For example, the animal simply pitches up the body to push over flimsy beams, whereas it often transitions (figure 1, orange) from pitching (figure 1, blue) to rolling (figure 1, red) into a gap and maneuver through stiff beams. A robophysical model was developed to study the principles of locomotor transitions [37]. It was propelled forward with a constant force and the body was oscillated to generate significant kinetic energy fluctuation, which emulated that in the cockroach body from self-propulsion when interacting with the obstacles [37]. A potential energy landscape model showed that the pitch and roll modes arose, as the system was attracted to distinct pitch and roll basins on a potential energy landscape [37]. Systematic experiments using the feedforward robot demostrated that locomotor transition from pitch to roll mode occurred when kinetic energy fluctuation from body oscillations exceeded the potential energy barrier between the pitch and roll basins [37]. However, the cockroach still made this transition even when kinetic energy fluctuation from body oscillation alone was insufficient [37]. A later study further found that the cockroach actively adjusted their body and appendages (e.g., head, abdomen, and legs) during obstacle traversal [31].

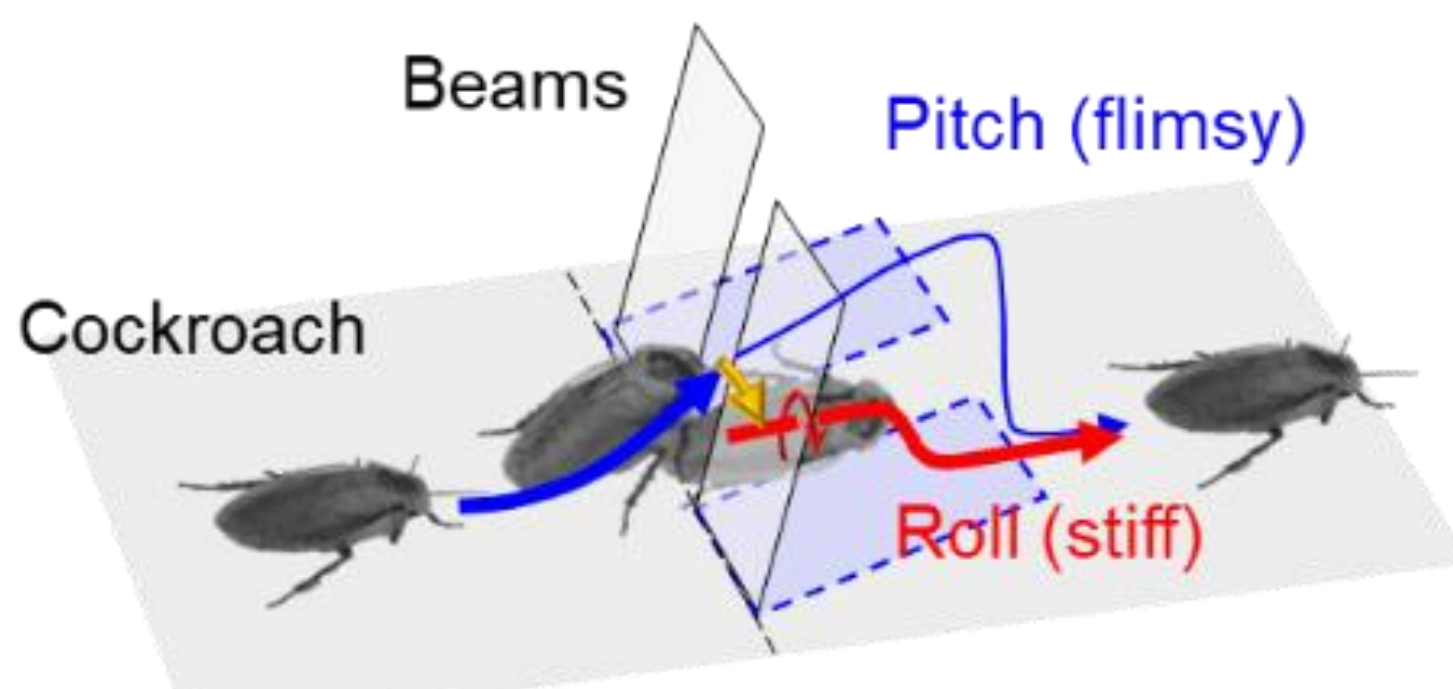

Figure 1. Cockroach beam traversal experiments show how cockroach traverses different beams. When beams are flimsy, cockroaches more likely use pitch mode (blue). When beams are stiff, cockroaches more likely use the roll mode (red). To traverse stiff beams, they often transition from the pitch to the roll mode (orange) [37].

Here, we hypothesized that environmental force sensing can help animals and robots traverse cluttered large obstacles, by guiding active adjustments to facilitate locomotor transitions. This was plausible because the cockroach used different traversal strategies when traversing beams of different stiffness: it pushed across flimsy beams but transitioned to rolling through the gaps when encountering stiff beams [37]. It is possible that the animal sensed the stiffness of obstacles during contact and used this information to guide their locomotor transitions to use the appropriate traversal strategy (see discussion in [31]). To test this hypothesis, we created a physics model and a simulation for a minimalistic robot, which can detect forces in its environment and actively switch between locomotor modes. We then used the model and simulation to study whether and how environmental force sensing helps traversal.

The minimalistic robot was modeled as an ellipsoidal body propelled forward, and the grass-like beams were modeled as two plates with torsion springs at their base on the ground, following the previous study [37]. The potential energy landscape model in the previous study did not fully model the dynamics of the system [37]. To address this, here we incorporated a contact force model, equations of motion, and force/torque controllers in the physics model. Using this model and force sensing, we developed a new traversal strategy, the force-feedback control, which estimated beam stiffness from contact forces and controlled the robot body to transition to the appropriate locomotor mode.

To demonstrate the advancement of the force-feedback strategy, we further built a multi-body dynamics simulation of the minimalistic robot interacting with the beams. We chose a simulation here because, compared to the simple physics model, the simulation implemented a more complex contact mechanics that modeled the collisions more accurately. Simulation results revealed that, compared to simply pushing across the beam obstacles or avoiding obstacle contact, the force-feedback strategy offered the robot a higher success rate of traversal, less energy consumed, and reduced the probabilities of becoming stuck in front of beams or flipping over. In addition, we demonstrated the robustness of beam stiffness estimation against body lateral and vertical oscillations, randomness in oscillation, and uncertainty in the robot's position sensing. We also found that a shorter sensorimotor delay reduced mechanical energy cost of traversal. Finally, we discuss the limitations of our method, its broader applications, and future work.

## 2. METHOD

### *1. Simulation robot design*

The simulation robot (figure 2A) built in Chrono Engine followed the design in our previous study [37]. It was modeled as a rigid ellipsoidal body similar to the discoid cockroach's body, with principal axes lengths $2a$ = 0.22 m, $2b$ = 0.16 m, and $2c$ = 0.06 m. An additional weight of 0.5 kg makes it bottom-heavy (the center of mass is lower than the geometric center by $h_c = 0.01$ m) to mimic the self-righting performance of the legged robot on the ground (figure 2A). The ellipsoidal body's total mass $M = 1$ kg and its moments of inertia along the three principal axes are $(I_1, I_2, I_3) = (1.0, 3.5, 5.0) \times 10^{-3}$ kg·m$^2$. The geometric center is $H = 0.105$ m from the ground initially.

The ellipsoidal body has three degrees of freedom in translation and two degrees of freedom in rotation (figure 2). The vertical rod hanging the ellipsoidal body is controlled by three motors to do vertical, lateral, and fore-aft motion. The center of the rotation overlaps with the geometric center. Similar to the previous design [37], the body has two rotation axes, which are the roll axis (an axis through the rotation center of the robot and parallel to the $X$-axis in the world frame; figure 2, red dashed line) and the pitch axis (the $y$-axis in the body frame; figure 2, blue dashed line.). Thus, the body's orientation can be represented by roll angle $\alpha$ and pitch angle $\beta$. We defined $\alpha = \beta = 0°$ at the equilibrium state. Two motors control the ellipsoidal body's rotation to mimic the cockroach's active pitch and roll adjustments.

Following our previous work, for simplicity and to facilitate energy landscape modeling, the grass-like beams were designed as rigid, vertical plates with torsion springs. The torque function of the spring is $\tau_s = k\theta - c_d\omega$, where $k$ is stiffness, $\theta$ is the relative beam deflection angle, $c_d$ is the coefficient of dampness ($c_d = 0.01\ N \cdot m \cdot s/rad$ in this paper), and $\omega$ is the relative deflection angular velocity. The grass is usually much lighter than the robot, so the mass of each beam is $m = 1$ g. The torsion springs align with the $Y$-axis and are symmetric about the $X$-$Z$ plane. The two beams are vertical to the ground when there is no deflection. The gap width between the two beams is $d = 0.138$ m, which is narrower than the width of the robot. Each beam's width, height, and thickness are $w = 0.04$ m, $L = 0.155$ m, and $t_h = 0.01$ m.

Chrono simulation requires material properties to calculate contact forces. In this paper, we set Young's modulus = $10^6$ Pa, Poison's ratio = 0.1, coefficient of restitution = 0, and coefficient of friction = 0.

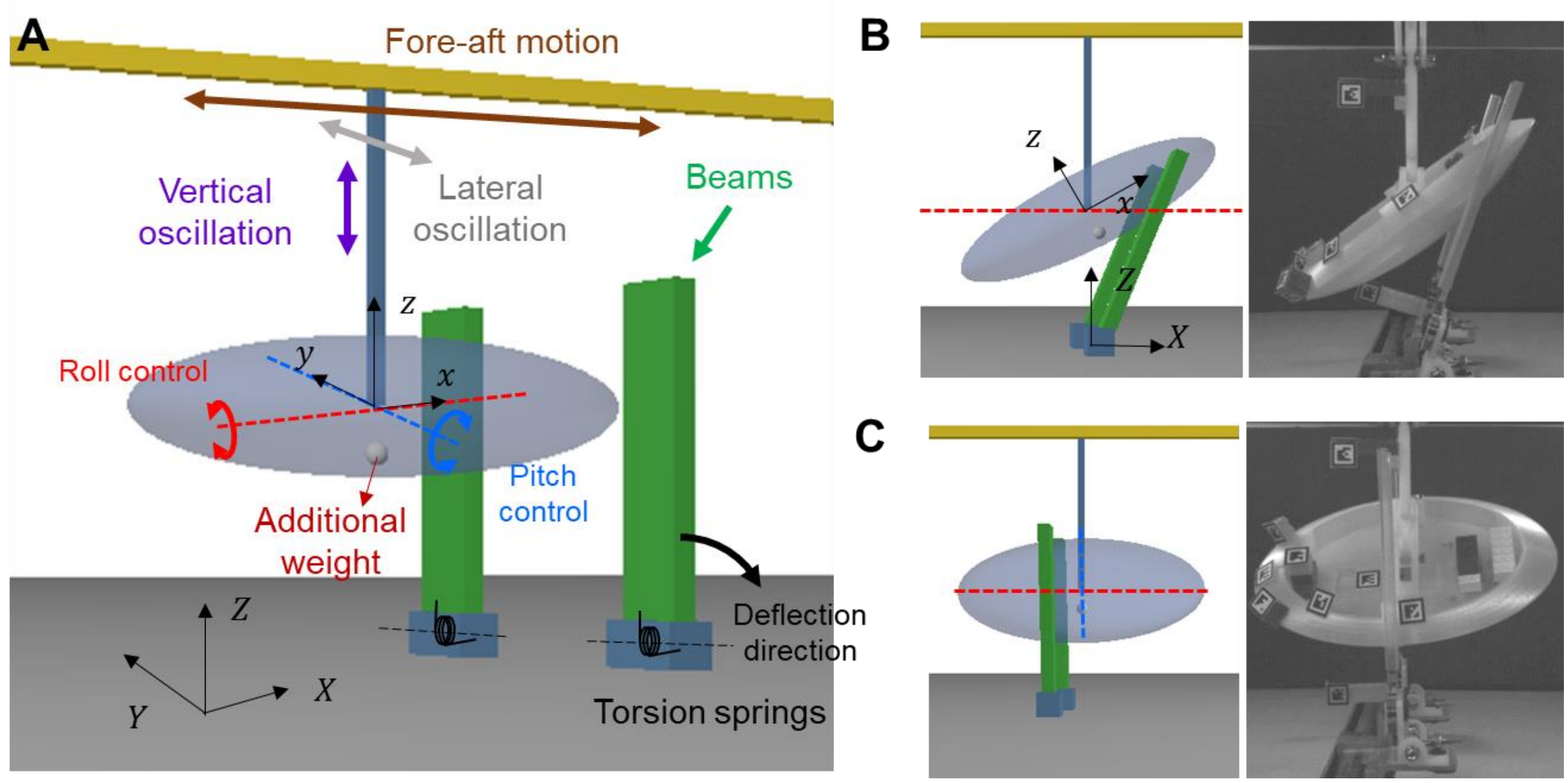

Figure 2. Multi-body dynamic simulation to beam traversal. (A) The simulation robot body has five degrees of freedom (fore-aft, vertical, lateral, roll (red arrow), and pitch (blue arrow)), which are controlled by motors. An additional weight makes it bottom-heavy. Each beam is connected to the ground via a torsion spring and can deflect in the vertical $X-Z$ plane. The world frame is $X-Y-Z$. Body frame is $x-y-z$, which is attached to the robot. (B-C) Snapshots of simulation and physical robot experiments from our previous study [37]. (B) Without vertical oscillation, both simulation and physical robots are stuck in pitch mode. (C) With sufficient vertical oscillation, both roll into the beam gap and traverse in the roll mode.

### 2. *Physics model*

We built a physics model (figure 3) based on the simulation robot design. The physics model includes the contact force model (figure 3c) to estimate beams' stiffness from force sensing, the potential energy landscape of the system, and motion planning and control to help the robot traverse beams.

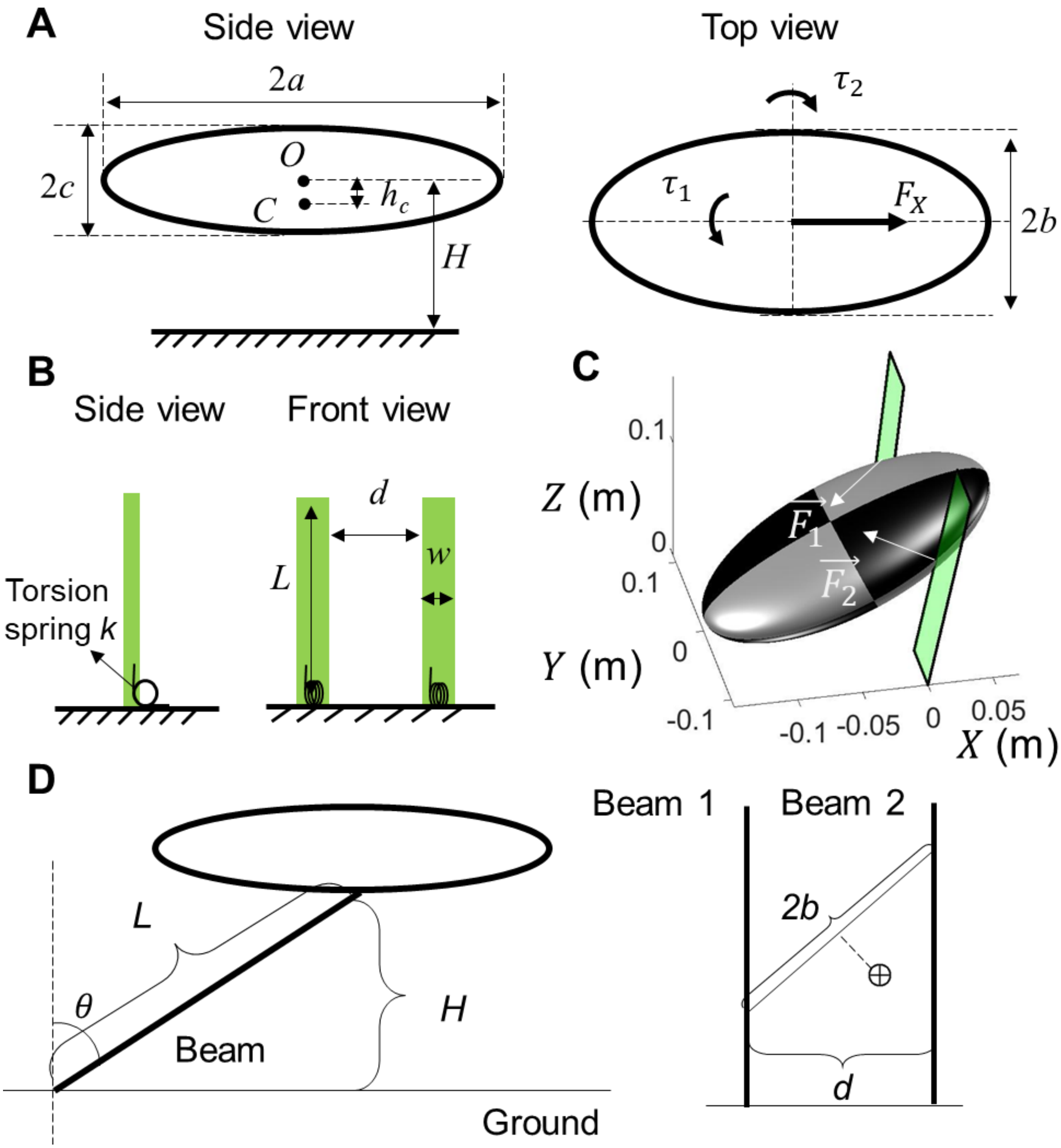

Figure 3. Physics model of robot-beam interaction. (A) Cockroach-inspired robot is modeled as an ellipsoidal rigid body. $F_X$ is propulsive force. $\tau_1$ and $\tau_2$ are roll and pitch torques along the roll and pitch axes in figure 2. The center of mass ($C$) is lower than the rotation center ($O$) by $h_c$. (B) Each beam is modeled as a rigid plate with a torsion spring. Stiffness is $k$. (C) $\vec{F}_1$ and $\vec{F}_2$ are theoretical contact forces with each beam calculated from the model. (D) Left: In pitch mode, the potential energy is approximately at a maximum when the bottom of the robot reaches the top of the beam. Right: In roll mode, the potential energy is approximately at a maximum when the robot rolls until there is no beam contact.

Besides the input force $F_x$ and torques ($\tau_1$, $\tau_2$) from controllers acting on the simulation robot, there are three external forces, which are the gravitational force $\vec{G}$ and contact forces $(\vec{F}_1, \vec{F}_2)$ with two beams (Fig. 2C). Given these, we can calculate the total force $\vec{F}$ and torque $\vec{\tau}$. Thus, according Newton–Euler equations, the dynamic equations of motion of the robot are:

$$\begin{pmatrix} \vec{F} \\ \vec{\tau} \end{pmatrix} = \begin{pmatrix} M & 0 \\ 0 & I \end{pmatrix} \begin{pmatrix} \vec{\alpha} \\ \vec{\beta} \end{pmatrix} + \begin{pmatrix} 0 \\ \vec{\omega} \times I\vec{\omega} \end{pmatrix} \tag{1}$$

where $M$ is the mass, $I$ is the moment of inertia about the center of mass, $\vec{\alpha}$ and $\vec{\beta}$ are acceleration and angular acceleration about the centor of mass, $\vec{\omega}$ is the angular velocity.

*3. Contact force model*

Because beams are much lighter than the robot, the inertial force of the beam is negligible compared to the contact forces. Therefore, we assumed that each beam maintains torque equilibrium during contact.

$$k_i\theta_i - c_{di}\omega_i + \vec{r}_{bi} \times \vec{F}_i \cdot \begin{pmatrix} 0 \\ 1 \\ 0 \end{pmatrix} = \frac{1}{2} mgL \sin\theta_i \tag{2}$$

where $k_i$ and $c_{di}$ are the stiffness and damping coefficient of the $i^{\text{th}}$ beam, $\theta_i$ and $\omega_i$ are its beam deflection angle and angular velocity, $\vec{r}_{bi}$ is the shortest displacement from the rotational axis of the $i^{\text{th}}$ beam to the point of application of $\vec{F}_i$.

Because the robot's body surface is smooth, the coefficient of friction is set to 0 in the simulation. Thus, the frictional force is negligible, and the contact force is normal to the tangent plane of the contact point on the ellipsoidal body's surface.

The normal direction of the ellipsoidal body's tangent plane at a point $(x, y, z)$ in the body frame is:

$$\vec{n} = (b^2c^2x, a^2c^2y, a^2b^2z) \tag{3}$$

Given the current position and orientation of the robot and Eqns. 2 and 3, we can calculate the theoretical contact forces $\vec{F}_1$ and $\vec{F}_2$ in the world frame.

*4. Estimate the stiffness from force sensing*

Forces measured by the force sensor are used to calculate the unknown parameters $k_1$ and $k_2$ in Eqn. (2) by minimizing the error between the theoretical contact force $\vec{F} = \vec{F}_1 + \vec{F}_2$ and the measured force data $\vec{F}_{sensed}$ (using the MATLAB 'fminsearch' function):

$$e = \sum_{i=1}^{N} \left| \vec{F}_{sensed}(t_i) - \vec{F}(k_1, k_2, t_i) \right| \quad (4)$$

where $N$ is the total number of the sampling force data points at the sampling time $t_i$. This process was repeated 100 times with different initial guesses of parameter values randomly selected, increasing the likelihood of finding a global minimum. In this paper, initial guess values of $k_{1,2}$ are selected from uniformly distributed random numbers in the range [0, 5] N·m/rad, because a beam with stiffness exceeding the range is like a rigid body and easy to be differentiated.

### 5. *Potential energy landscape*

There are two different locomotor modes to traverse stiff and flimsy beams: the pitch and roll modes [37] (figure 1). To determine a less energetically-costly mode, we calculated the potential energy barrier during the entire traversal. The potential energy of the system consists of the gravitational potential energy of the robot and the elastic potential energy of the beams:

$$E(X, \alpha, \beta) = MgZ_{CoM}(\alpha, \beta) + \frac{1}{2}mgL\,(\cos\theta_1 + \cos\theta_2) + \frac{1}{2}k_1{\theta_1}^2 + \frac{1}{2}k_2{\theta_2}^2 \quad (5)$$

where $Z_{CoM}(\alpha, \beta)$ is the height of the center of mass with respect to the ground, $\theta_1$ and $\theta_2$ are beam deflection angles, that are functions of α, β, and $X$.

We varied $X$ from –0.12 m to 0.12 m with an increment $\Delta X$ = 0.002 m and α and β from –90° to 90° with an increment Δα (Δβ) = 2° to obtain a 3-D potential energy landscape over the entire (α, β, $X$) workspace during traversal (figure 4A).

### 6. *The choice of locomotor mode depends on beam stiffness*

Using the potential energy landscape, we calculated the critical stiffness for the choice of the pitch and roll modes based on the potential energy barrier. For simplicity, we assumed that the two beams had the same stiffness ($k_1 = k_2 = k_0$) to estimate the potential energy barrier using each mode (this assumption only applies to this section). For the pitch mode, we assume that the potential energy reaches the maximum when the robot's bottom reaches the top of the beams where beams are deflected most (figure 3D, Left). Thus, its potential energy barrier is $PE_{pitch} \sim k \cdot \arccos\,(H,\ L)^2$. For the roll mode, the potential energy reaches the maximum when the robot rolls until there is no contact in the entire traversal (figure 3D, Right). Thus, its potential energy barrier is $PE_{roll} \sim Mgh_c$. Because $PE_{roll}$ does not depend on beam stiffness whereas $PE_{pitch}$ is proportional to it, the barrier difference $PE_{pitch} - PE_{roll}$ monotonically increases with $k$. Although the traversal is dynamic and one needs to consider kinetic energy and inertia, considering its complexity, we can use this quasi-static analysis as a quick alternative way for determing which locomotor mode costs less energy for a given stiffness of beams. Roughly speaking, when the beam stiffness is above a threshold $k_0 = \frac{Mg\, h_c}{\arccos{(H,L)^2}}$, the roll mode is better because it overcomes a lower barrier; below it, the pitch mode overcomes a lower barrier.

*7. Motion planning*

To further obtain the planned trajectory in *X*, roll, and pitch space, the mechanical energy cost of each trajectory is defined to measure the difficulty of traversal. A directed graph is created based on the energy landscape built in Sec. 2.5 (figure 4B). We defined the states in the energy landscape as vertices (Sec. 2.5) and the connections between vertices as directed edges. Each vertex connects with its six closest neighbors, which are $(X \pm \Delta X, \alpha, \beta)$, $(X, \alpha \pm \Delta\alpha, \beta)$, and $(X, \alpha, \beta \pm \Delta\beta)$. A path in the network is composed of vertices and directed edges. Thus, a path's total mechanical energy cost is the sum of all directed edges within it. The actuator needs to do work for the robot to move to a state with higher potential energy along a directed edge. Thus, the cost of this directed edge is positive and defined as this potential energy increases. The decrease in potential energy can transform into kinetic energy for the robot to move to a state with lower potential energy along a directed edge. In this case, we assumed that the actuators do no work and that the cost of the directed edge is zero (figure 4B).

For each trajectory, we defined the initial state as the state when the active control began after contact with beams and a sensorimotor delay. The target state was defined to be the state with the horizontal pose ($\alpha = \beta = 0°$) after traversal ($X = 0.1$ m). The sensorimotor delay $\Delta t$ is defined as the time interval from the instant of first contact with beams $t_c$ to the instance when active control began $t_s$, i.e., $\Delta t = t_s - t_c$. Given the directed weighed graph (figure 4C), the A-star algorithm [41] is used to search for the path of the minimal cost between the initial and target states, with the heuristic function equals to the potential energy difference of two states. We took this path as the planned path.

*8. Motion control*

After motion planning, we controlled the simulation robot to track the planned trajectory, which was the planned path with a constant forward speed of $v_X = 0.05$ m/s. Based on the position error, we used a model-based feedback control to obtain the control input at every 0.002 s:

$$\vec{u} = F_{ex}(\vec{q}) - K_p(\vec{q} - \overrightarrow{q_d}) \tag{6}$$

where $\vec{q} = (X, \alpha, \beta)$ is the current state, and $\overrightarrow{q_d} = (X_d, \alpha_d, \beta_d)$ is the desired state. $K_p$ are positive feedback gains (we chose 0.1 in this paper). $\vec{u} = (\tau_1, \tau_2, F_X)$ is the control input (figure 3B). $F_{ex}(\vec{q})$ is the external force calculated from the contact force model (Sec. 2.3), including contact forces with beams and the gravitational force. With the control input $\vec{u}$, the current state is controlled to approach the desired state at every time step.

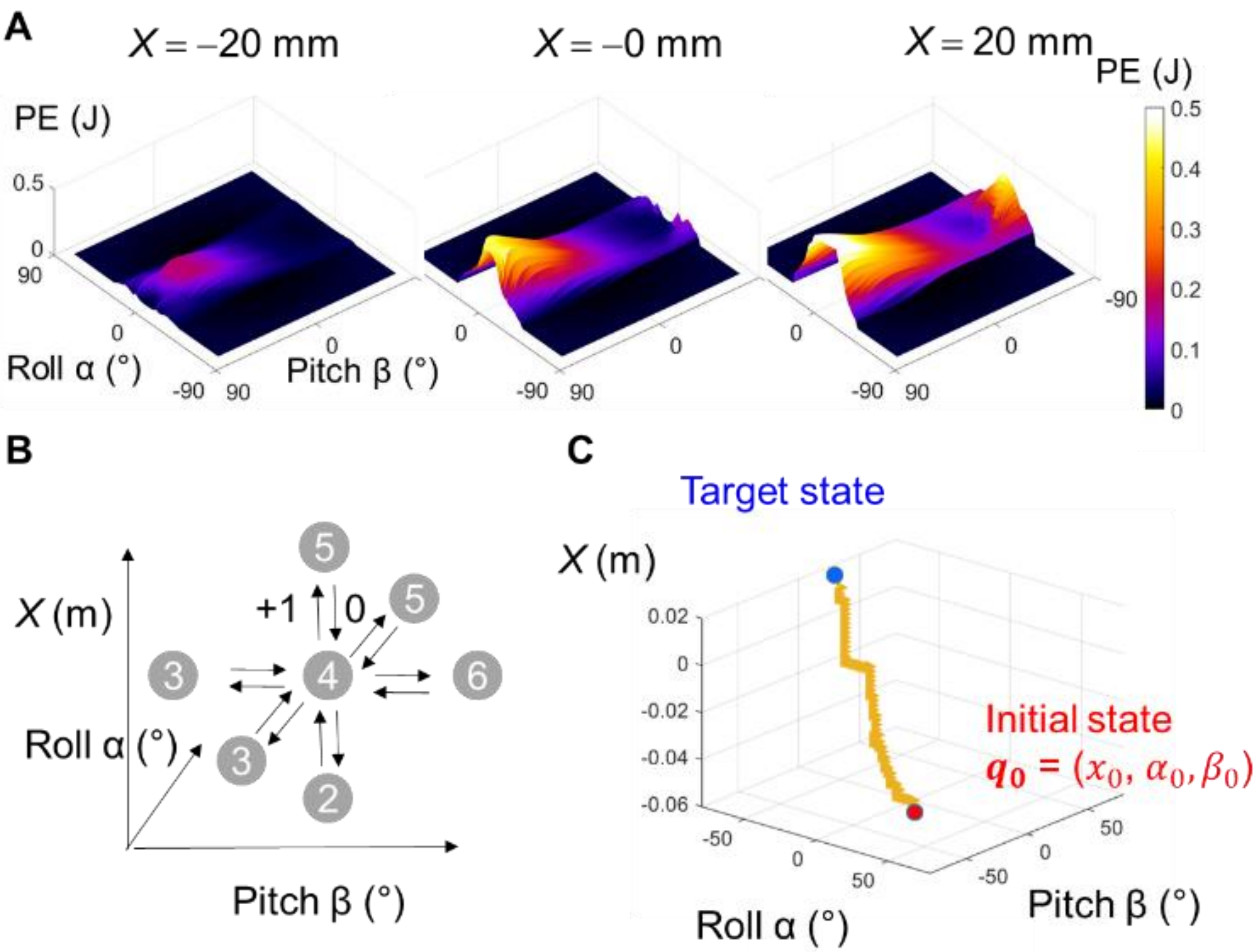

Figure 4. Motion planning. (A) 3-D potential energy landscape of the robot-beam system. (B) Schematic of the directed graph based on the potential energy landscape. Circles are states. Values in circles show representative potential energy at each state. (C) Minimal cost path from the initial state to the target state in the potential energy landscape.

## 3. RESULTS

### *1. Traversing flimsy and stiff beams without force sensing*

We first studied traversal without force sensing. The stiffness of the beams was selected based on the estimated critical stiffness in Sec. 2.6 so that one group of beams is stiff (easier to traverse using roll mode) and the other group is flimsy (easier for pitch mode). Substituting the parameters' values given in the simulation robot design (Sec. 2.1) to the function of the critical stiffness $k_0 = \frac{Mg\,h_c}{\arccos(H,L)^2}$ (Sec. 2.6), we obtained $k_0$ = 0.146 N·m/rad. Thus, we chose $k_{low}$ = 0.01 N·m/rad and $k_{high}$ = 0.2 N·m/rad as the stiffness of flimsy and stiff beams, respectively.

To quantify the difficulty of traversal, we used the energy cost, defined as the work done by actuators in the entire traversal:

$$E_{cost} = \sum_{i=1}^{m}[(F_X v_X)^+ + (\tau_1\omega_1)^+ + (\tau_2\omega_2)^+]\,dt \quad (7)$$

where $m$ was the number of control steps in the entire traversal time, $dt$ = 0.002 s was the control time step (Sec. 2.8). $(F_X v_X)$ was the power of the force $F_X$. $(\tau_1\omega_1)$ and $(\tau_2\omega_2)$ were the powers of two torques $(\tau_1, \tau_2)$. The symbol "+" meant $(f)^+ = \max\{f, 0\}$. We assumed that actuators would not restore energy.

*i. Feedforward control without force limitation*

For this control strategy, we simulated the robot traversing flimsy and stiff beams with only propulsive force but without roll and pitch control.

We first ran simulation trials to traverse flimsy and stiff beams without limitation in the propulsive force. The actuator controlled the robot to move forward with a constant speed of $v_X$= 0.05 m/s. When encountering flimsy beams, the robot pushed down beams with a small body pitch angle (< 4°) (figure 5A, bottom; Supplementary Material) and small contact force (< 0.13 N). Its energy cost was 14.8 mJ. When encountering stiff beams, the robot was pushed to pitch up and down over a wide range of angles ([−37°, 38°]) (figure 5A, top; Supplementary Material), and the maximum resistive force reached 1.5 N. The large range of pitch rotation could easily make robots lose stability and flip over. In addition, its energy cost was 185.2 mJ, which was much larger than the flimsy beam traversal.

*ii. Feedforward control with force limitation*

A strong propulsive force can propel the model through the beams. However, in reality, both animals and robots have limited force/torque output. To avoid tremendous forces/torques input, we set limits for them ($|F_x| \leq 1$

N, $|\tau_1| \leq 0.1$ N·m, $|\tau_2| \leq 0.1$ N·m) as an example. When any component of $\vec{u}$ calculated from Eqn. 6 exceeds limits, this component is set to be the maximum or minimum values within the limit. We then ran simulation trials with force/torque limitations. When encountering flimsy beams, the robot behaved the same as the trials with unlimited force because the resistive force was smaller than the force limit (figure 5B, bottom; Supplementary Material). However, when encountering stiff beams, the robot was stuck in front of the beams (figure 5B, top; Supplementary Material).

Thus, an open-loop robot that can only move forward can work well when encountering flimsy obstacles. However, it will behave poorly when encountering stiff obstacles, such as losing stability, high energy cost, and being trapped forever.

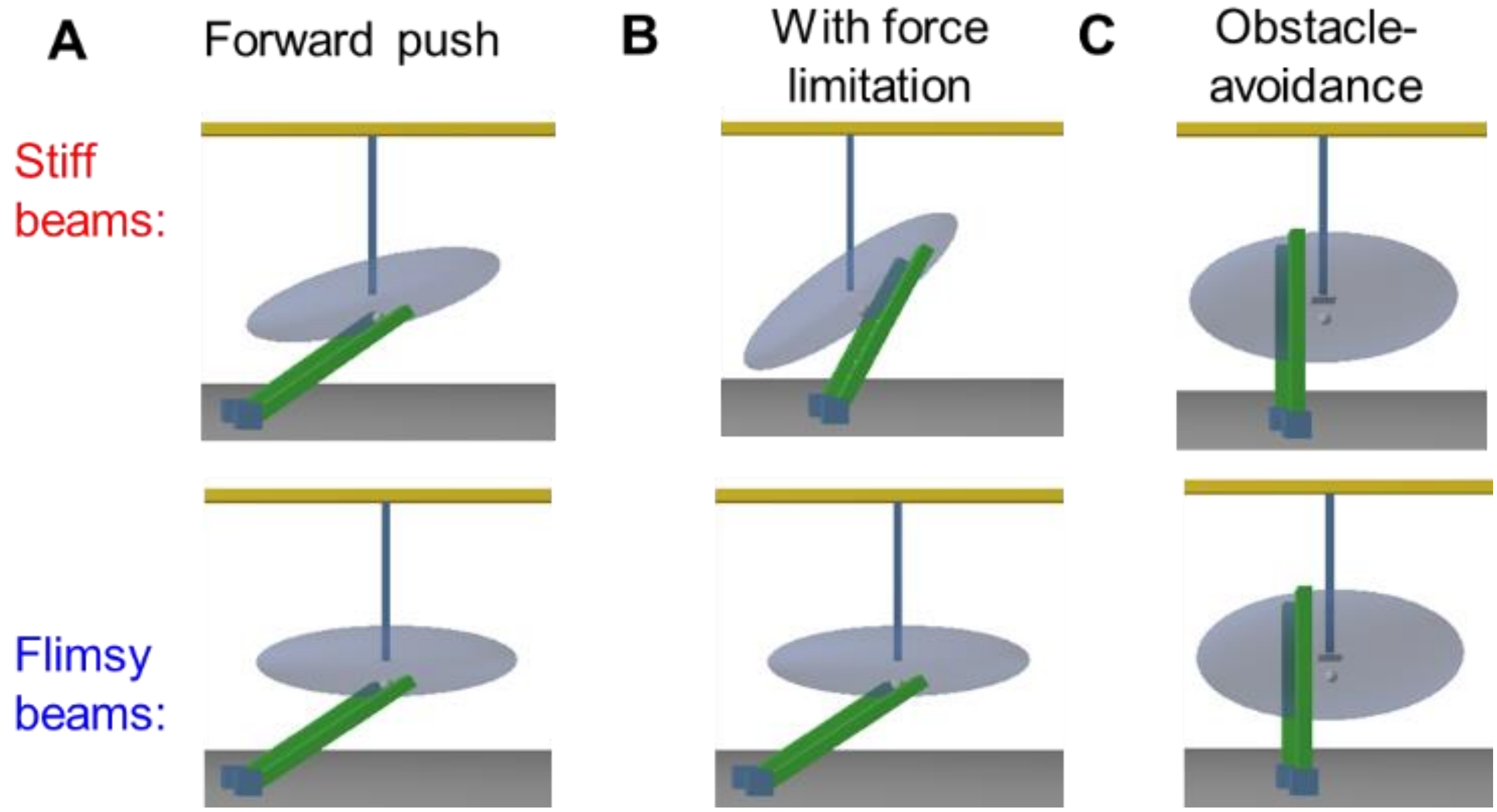

Figure 5. Representative snapshots of robot traversing stiff and flimsy beams in simulation based on different strategies. (A) Feedforward pushing strategy with unlimited propulsive force. Both traversed beams. (B) Feedforward pushing strategy with limited propulsive force. It was stuck on stiff beams and succeeded in traversing flimsy beams. (C) Obstacle-avoidance strategy. Both succeeded in rolling to traverse beams.

*iii. Always avoid obstacles*

Another control method without force sensing is to avoid obstacles, which has been widely used in vehicles and mobile robots [8,9]. The robot used the roll motor in the simulation to roll through the gap. First, we used the potential energy landscape to obtain the minimal roll angle that the robot could traverse with the beams' elastic energy keeping zero. In other words, the robot could traverse without contact with beams. Then, we applied feedback control (Eqn. 6) to reach the desired roll angle before traversing and recover it to its original orientation after traversing.

This obstacle-avoidance approach was based on vision, so the robot behaved the same when traversing flimsy and stiff beams (figure 5C; Supplementary Material). Its energy cost was 35.8 mJ. We noticed that this approach saved much energy compared to the forward-pushing approach (185.2 mJ) when the obstacle was stiff. However, the obstacle-avoidance approach was not the best option for the flimsy beams. Its energy cost was higher than the forward-pushing approach (14.8 mJ), and its large body roll angle increased the difficulty of maintaining the robot's stability for the real animals/robots.

In conclusion, two traversal approaches without force sensing have advantages and disadvantages. A new approach with force sensing can combine the advantages of the previous two and help the robot better traverse beams with varying stiffness, which we will show in the Sec. 3.2 and 3.3.

*2. Estimating beam properties in simulation*

We can know the position and orientation of the robot relative to obstacles via visual sensing, and their contact force via force sensing. In addition, we can use the contact model to estimate the beam stiffness (Sec. 2.4).

*i. Estimation of beam stiffness under random body oscillations*

Inspired by the locomotion of cockroaches and legged robots, whose self-propulsion by legs triggered substantial body oscillation, we induced lateral and vertical oscillations into the motion (figure 2). Then, we studied the impact of oscillations to test the robustness of the approach.

Motors applied position control to control the body's motion. The lateral oscillation followed the triangle wave function, with the frequency $f$ and amplitude $A_l$. The vertical oscillation followed a sum of 30 sine functions to induce randomness. The amplitude of each sine function is randomly picked from a uniform distribution from –0.5 mm to 0.5 mm. The frequencies of 30 sine functions are $\frac{f}{50}, \frac{2f}{50}, \frac{3f}{50}, \ldots, \frac{30f}{50}$. Randomness was induced into the vertical oscillation because we wanted to let the result more general as statistical results.

In the simulation, the robot moved forward with a constant forward velocity $v_x = 0.05$ m/s while oscillating in the lateral and vertical directions. The sum of contact forces ($\vec{F}_{sensed}$) between the robot and two beams were sensed with a sampling rate of 40 Hz.

Using the sensed force data, we used the estimation method to obtain the estimated values of beam stiffness. First, we varied the lateral oscillation amplitude ($A_l$ = 1, 2, 3 mm) and frequency ($f$ = 2, 4, 6 Hz). Then, we ran five trials for each to use the average value as its statistical result.

To quantify the estimation accuracy, we calculated the relative error of stiffness, which was the ratio of the estimated error and its true value $e_{ki} = \frac{|\widehat{k_i} - k_i|}{k_i}$, where $\widehat{k_i}$ is the estimate value ($i = 1, 2$). $k_i$ is the true value of stiffness, and $k_1$ = 0.5 and $k_2$ = 0.5 N·m/rad in this section (Sec. 3.2).

We recorded the estimation results ($\widehat{k_1}$, $\widehat{k_2}$) for two beams with varying oscillation frequencies (Table 1, oscillation amplitude = 1 mm) and oscillation amplitudes (Table 2, oscillation frequency = 2 Hz) during forwarding motion.

Although the oscillation caused more collisions and intermittent contact, this approach still obtained an accurate estimation (relative error $e_k$ < 5% in most trials), which was a prerequisite for motion planning and control.

Table 1. Stiffness estimation (N·m/rad) with varying oscillation frequencies.

| Frequency (Hz) | 2 | 4 | 6 |
|---|---|---|---|
| $\widehat{k_1}$ ($e_{k1}$) | 0.484 (3.3%) | 0.475 (5.1%) | 0.477 (4.7%) |
| $\widehat{k_2}$ ($e_{k2}$) | 0.473 (5.4%) | 0.479 (4.3%) | 0.475 (4.9%) |

Table 2. Estimation of stiffness (N·m/rad) with varying oscillation amplitudes.

| Amplitude (mm) | 1 | 2 | 3 |
|---|---|---|---|
| $\widehat{k_1}$ ($e_{k1}$) | 0.484 (3.3%) | 0.482 (3.6%) | 0.483 (3.3%) |
| $\widehat{k_2}$ ($e_{k2}$) | 0.473 (5.4%) | 0.477 (4.6%) | 0.477 (4.6%) |

*ii. Estimation of beam stiffness with position sensing error*

Although oscillation and randomness were induced in the above estimation, we still obtained the accurate relative position between the robot and the obstacles. However, in practice, the robot's position sensing may not be accurate because of its intense oscillations.

To study the impact of uncertainty in position on the estimation accuracy, we used an inaccurate position sensing in $Y$, by assuming the value $Y = 0$ mm during traversal regardless of lateral oscillation. Thus, the larger the lateral oscillation amplitude, the more significant inaccuracy of the $Y$ position was.

We ran simulations with an oscillation frequency of 6 Hz and varied the oscillation amplitude = 1, 2, and 3 mm. For each oscillation amplitude, we ran five trials. We plotted the mean of relative errors ($\frac{e_{k1}+e_{k2}}{2}$) of beams' stiffness as a function of oscillation amplitude (figure 6). From the trend on the plot, we found that the oscillation amplitude increased the relative error of estimation in general. In addition, the variance of estimation among different trials also increased when increasing oscillation amplitudes, because the amount of uncertainty increased. Despite the uncertainty in position increased estimation error, the estimation accuracy was still acceptable (relative error < 15%), which could differentiate between stiff and flimsy beams after contact. It represented the robustness of the estimation against the error in position sensing.

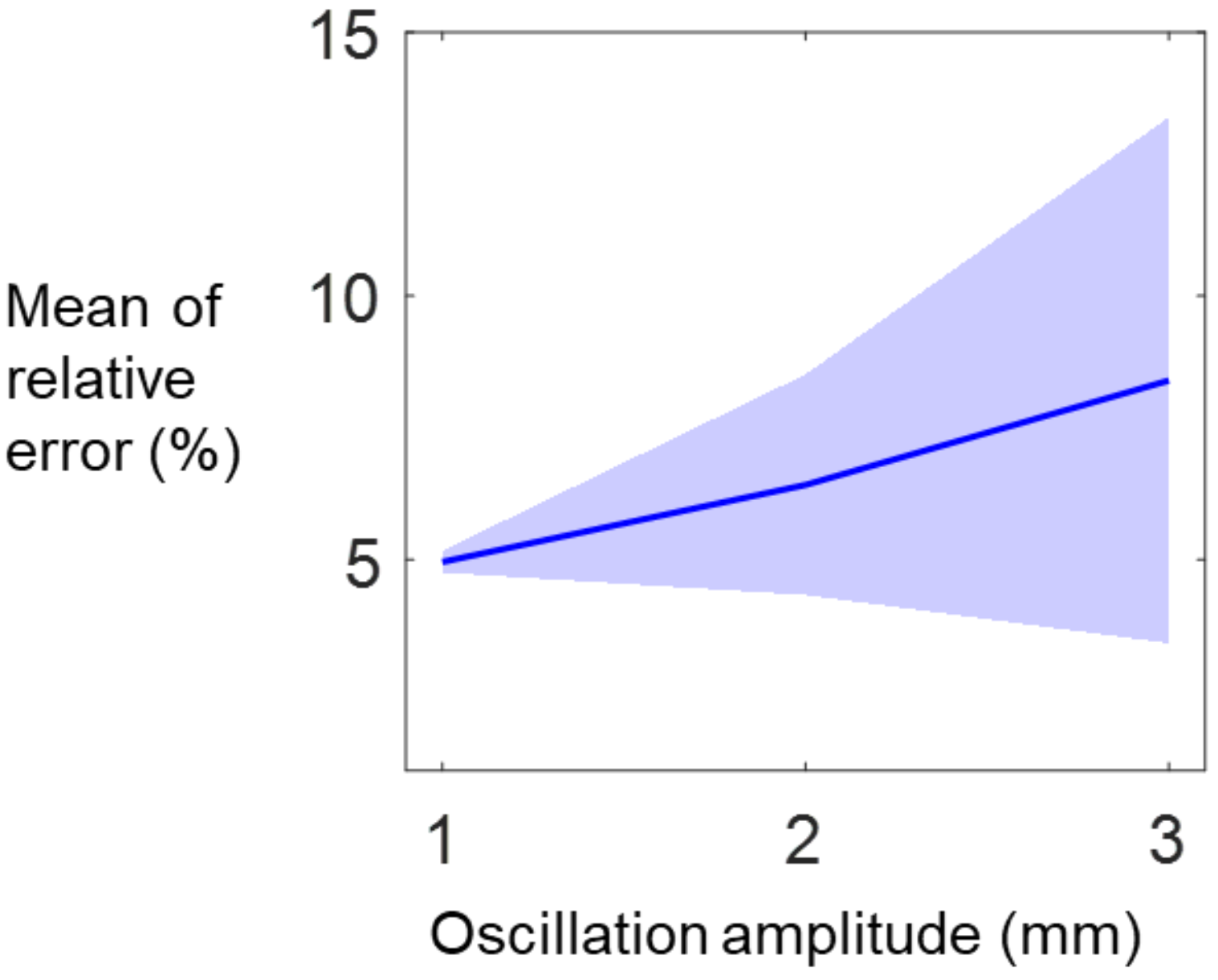

Figure 6. Mean of relative estimation error ($\frac{e_{k1}+e_{k2}}{2}$) as a function of oscillation amplitude (five trials).

*iii. The affect of the sensing time on estimation*

Given our constant forward speed, it only took the robot 4.4 seconds to traverse a distance equal to its body length. Thus, the robot needs to react fast. Otherwise, it cannot roll its body through the gap of stiff beams in time and will cost extra energy and lose stability (Sec. 3.1.1). Thus, the time cost of our method is important.

An inevitable time cost is sensing time, which is required to obtain data points to estimate stiffness. Therefore, we studied how estimation accuracy depended on the sensing time to know how long it should be to obtain a reasonable estimate. We tested sensing time $T_s$ = 25, 50, 100, and 200 ms with a sampling rate of 40 Hz. A longer sensing time means more data points can be used in model fitting. As expected, sensing time increased estimation accuracy (Table 3). However, for sufficiently large numbers of data points (sensing time ≥ 50 ms), more data points would not improve estimation accuracy substantially, because the randomness and modeling error became dominant (Table 3).

Table 3. Estimation based on different sensing times ($f$ = 2 Hz, $A_1$ = 1 mm).

| $T_s$ | 25 ms | 50 ms | 100 ms | 200 ms |
|---|---|---|---|---|
| $\widehat{k_1}(e_{k1})$ | 0.531 (6.2%) | 0.513 (2.6% ) | 0.516 (3.2%) | 0.522 (4.4%) |
| $\widehat{k_2}(e_{k2})$ | 0.383 (23.4%) | 0.473 (5.4%) | 0.491 (1.8%) | 0.488 (3.2%) |

*3. Traversing beams with the force-feedback control strategy*

We finally simulated with the force-feedback control strategy to study how they would help the robot traverse flimsy and stiff beams. After the initial contact, we obtained the force data (sampling during $T_s$ = 100 ms) to estimate the beam stiffness. Then, we constructed a potential energy landscape and performed motion planning. Finally, we controlled the robot to track planned trajectories to traverse two kinds of beams.

The simulation showed that the control method worked well and the traversal was successful in both cases (figure 7; Supplementary Material). The mechanical energy cost to traverse flimsy beams was 15.0 mJ, which was almost the same as the forward pushing strategy (14.8 mJ; Sec. 3.1.1). The energy cost to traverse stiff beams was 17.4 mJ, which was better than the previous two strategies without force sensing (185.2 mJ and 35.8 mJ; Sec. 3.1). It was even more energy efficient than the obstacle avoidance method, because the robot could lean against the beams between the gap and the interaction forces with beams helped the robot to roll.

In addition, the traversal strategy based on force sensing and optimization using energy landscape generated the same traversal modes to traverse beams when encountering the flimsy and stiff beams, as observed in cockroaches (figure 7 versus figure 1). For flimsy beams ($k_{low}$ = 0.01 N·m/rad), the ellipsoidal body pitched up and moved over beams. For stiff beams ($k_{high}$ = 0.2 N·m/rad), the body rolled through the gap.

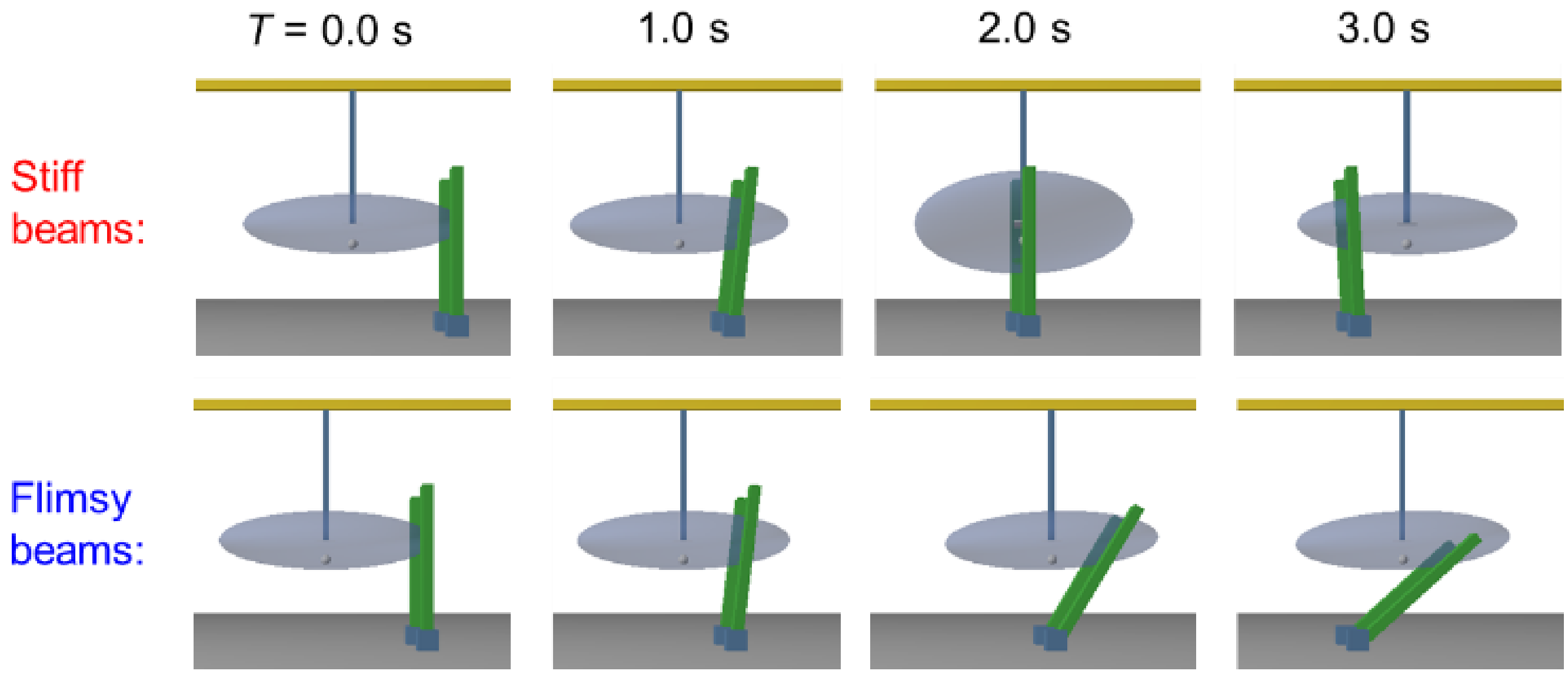

Figure 7. Robot traversed stiff and flimsy beams under motion control based on force sensing.

*4. Traversal performance with different sensorimotor delay*

The sensorimotor delay is evitable in both animals and robots. Here, we used our simulation to study the influence of sensorimotor delay on traversal performance. The sensing time $T_s$ described above is also a component of sensorimotor delay. We tested three different sensorimotor delays $\Delta t$ = 320, 480, and 640 ms in simulation. In addition, we calculated the energy cost from actuators (Eqn. 7) for trials of different sensorimotor delays.

The longer the sensorimotor delay was, the more the robot pitched up when it started active control (figure 8A). Despite this, the robot successfully traversed in all trials (Supplementary Material). Furthermore, we found that the longer the sensorimotor delay, the more energy cost from actuators (figure 8B). The faster the robot could estimate the properties of obstacles and respond after encountering obstacles, the more energy it would save.

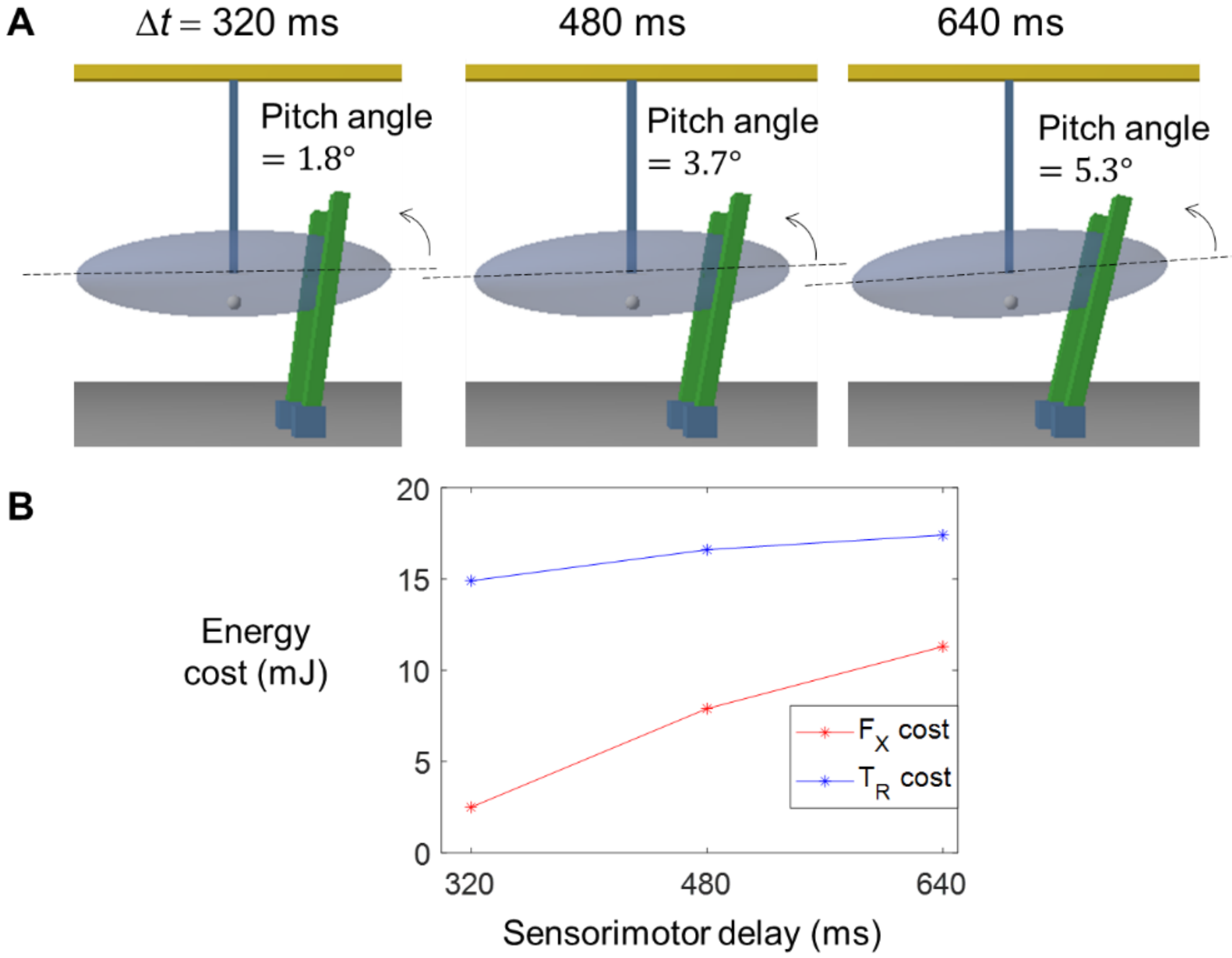

Figure 8. Influence of sensorimotor delay. (A) States after different sensorimotor delays. (B) In different sensorimotor delays, energy cost from the forwarding motor ($F_X$ cost) and rotation motors ($T_R$ cost).

## 4. Discussion

The theoretical model and simulation showed that environmental force sensing can determine previously unknown beam properties and inform planning and control accordingly to traverse cluttered obstacles with physical interaction. We compared this new traversal strategy, force-feedback strategy, with two common traversal strategies, forward pushing and obstacle avoidance, in simulation. The force-feedback strategy based on force sensing made the robot's obstacle traversal more energy-efficient and prevented it from becoming stuck in front of beams or flipping over.

Force/torque sensing helps mobile robots to obtain surface properties of ground [42–44], detect the contact with obstacles [45], and correct localization errors [46]. Our modeling study demonstrated how contact force sensing can also help estimate terrain resistance (stiffness of beam obstacles) and allow robots to better traverse cluttered large obstacles with increased performance. This fundamental understanding from our modeling work will be directly be useful for RHex-class robots [47,48] moving in complex 3-D terrain, and it will also catalyze more diverse robot platforms to sense and use body-obstacle physical interaction to better move in the real world [49–51].

In addition to robotics, our work also has implications for biology. The discoid cockroach (and other animals that rely on physical interaction to traverse cluttered large obstacles) may be able to estimate obstacle physical properties through force sensing (e.g., using load-sensitive mechanoreceptors [52]). Our modeling showed that knowing these properties may help them use locomotor modes that are less costly to traverse. Such mechanical sensory feedback can complement their eyes [53] and antennae [54] to sense the visual/geometric information environment and help them better traversing cluttered obstacles with reduced energetic cost, which is important in animal locomotion [55].

We are currently developing a robophysical model of the body-beam interaction system, instrumented with custom internal 3-D force sensors and distributed external contact sensors [56]. Building on the modeling insight here, we will use it to further study the principles of mechanical sensory feedback control and demonstrate advancement in a real robotic system.

Our work has limitations that require further examination. We neglected the effect of body inertia in motion planning and assumed constant forward speed for the robot, which likely limited its traversal performance. To overcome these limitations, we attempted to utilize optimal control. The optimal trajectory can be obtained by minimizing the cost function of the optimal control (Eqn. 7) using a physics model. However, these efforts are limited by the non-smooth nature of contact force and complexity of the dynamics model. At present, optimal control techniques do not effectively address these complexities.

For simplicity, we modeled obstacles as rigid plates with torsion springs, but our method may in principle also be applied to more complex obstacles, such as cantilever beams that can deflect at any point, an elastic rod that can deflect in all directions, or movable rigid bodiess such as rock or boxes. Using detected contact forces, our method should be able to estimate not only unknown parameters in obstacles, but also uncertain states of robots, such as position and orientation. However, the accuracy of estimation will degrade as the number of unknown parameters increases.

The predictive physics model not only helps estimate unknown parameters but also helps in control. For example, the feedback control (Eqn. 6) requires real-time reading of the robot's states and the external forces.

However, because there is a time delay in sensors, the actual control input is calculated using the state and external forces in the previous moment. Therefore, when velocities or accelerations are large, the control may not be timely and may lead to unexpected collisions. In this case, the physics model can predict the state and external forces, which can help better control timely.

## ACKNOWLEDGMENTS

We thank Ratan Othayoth, Yaqing Wang, and Qiyuan Fu for discussion. This work was supported by an Arnold & Mabel Beckman Foundation Beckman Young Investigator Award, a Burroughs Wellcome Fund Career Award at the Scientific Interface to C.L. Author contributions: Q.X. and C.L. conceived the study; Q.X. designed the study, developed the theoretical model, performed simulation, and analyzed data; and Q.X. and C.L. wrote the paper.